\documentclass[letterpaper, 10 pt, conference]{ieeeconf}  

\IEEEoverridecommandlockouts                              

\usepackage{amsmath} 
\usepackage{amssymb}  
\usepackage{graphicx}
\usepackage{makecell} 
\usepackage{multirow}
\usepackage{multicol}
\usepackage{cite}

\usepackage{booktabs}
\renewcommand{\baselinestretch}{0.97}

\title{\LARGE \bf
A Contact-Driven Framework for Manipulating in the Blind
}

\author{Muhammad Suhail Saleem, Lai Yuan, and Maxim Likhachev
\thanks{All authors are affiliated with The Robotics Institute, Carnegie Mellon University, Pittsburgh, PA 15213, USA. {\small e-mail: \tt \{msaleem2, laiy, mlikhach\}@andrew.cmu.edu}
}
}

\begin{document}

\maketitle
\thispagestyle{empty}
\pagestyle{empty}

\begin{abstract}
Robots often face manipulation tasks in environments where vision is inadequate due to clutter, occlusions, or poor lighting—for example, reaching a shutoff valve at the back of a sink cabinet or locating a light switch above a crowded shelf. In such settings, robots, much like humans, must rely on contact feedback to distinguish free from occupied space and navigate around obstacles. Many of these environments often exhibit strong structural priors—for instance, pipes often span across sink cabinets—that can be exploited to anticipate unseen structure and avoid unnecessary collisions. We present a theoretically complete and empirically efficient framework for manipulation in the blind that integrates contact feedback with structural priors to enable robust operation in unknown environments. The framework comprises three tightly coupled components: (i) a contact detection and localization module that utilizes joint torque sensing with a contact particle filter to detect and localize contacts, (ii) an occupancy estimation module that uses the history of contact observations to build a partial occupancy map of the workspace and extrapolate it into unexplored regions with learned predictors, and (iii) a planning module that accounts for the fact that contact localization estimates and occupancy predictions can be noisy, computing paths that avoid collisions and complete tasks efficiently without eliminating feasible solutions.  We evaluate the system in simulation and in the real world on a UR10e manipulator across two domestic tasks—(i) manipulating a valve under a kitchen sink surrounded by pipes and (ii) retrieving a target object from a cluttered shelf. Results show that the framework reliably solves these tasks, achieving up to a $2\times$ reduction in task completion time compared to baselines, with ablations confirming the contribution of each module.
\end{abstract}


\section{INTRODUCTION}

\begin{figure}
\centering
\includegraphics[width=0.7\linewidth]{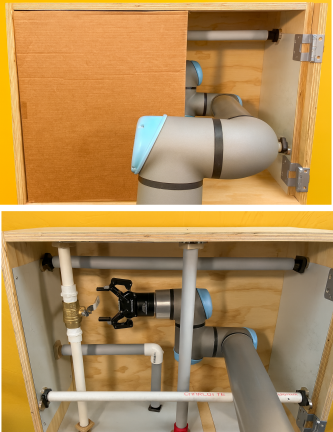} 
\caption{A robot reaching a shutoff valve at the back of a cluttered cabinet. As visual sensing is occluded by the shelf door (top), the robot relies on contact feedback to navigate around obstacles.}  \label{Fig: Intro image}
\vspace{-1em} \vspace{-0.2cm}
\end{figure}

Robots frequently encounter manipulation tasks in environments where visual sensing is inadequate. Consider the scenario in Fig.~\ref{Fig: Intro image}, where a robot is tasked with reaching a shutoff valve inside a cabinet under a kitchen sink. Occlusions, inconvenient positioning, and lack of direct line-of-sight make it difficult to visually inspect the region around the valve. Much like humans, the robot must instead rely on contact feedback to maneuver around obstacles and reach its target. As it moves through the cavity, it inevitably collides with pipes or brackets. By reasoning about where these contacts occur and combining this information with prior knowledge of typical obstacle structures encountered in the environment—for example, pipes are commonly found in this space and they often span the entire width of the cabinet—the robot can iteratively refine its understanding of free space and adjust its trajectory to the goal. 

This example illustrates a broader class of manipulation problems where robots must leverage contact to operate in unknown environments: retrieving an item from the back of a cluttered shelf, locating a switch hidden behind clutter, or reaching through scaffolding on a construction site. In such domains, strong priors about typical obstacle structures exist, but the specific obstacles in any given scene—and their precise placements—remain uncertain. 

To this end, we propose a theoretically complete and empirically efficient framework that integrates contact feedback with structural priors to enable robust manipulation in uncertain environments. Unlike much of the prior work on contact-rich manipulation, which has focused on high-resolution tactile sensing at the end-effector \cite{schneider2009object} \cite{comi2024touchsdf}, our approach addresses the fact that contacts in cluttered settings can occur anywhere along the arm. Full-body tactile skins \cite{skin1} that provide such coverage remain costly and fragile, whereas joint positions and torques are widely available on modern manipulators and offer a noisy but accessible alternative. We therefore leverage torque-based contact feedback as the primary sensing modality, augmented by learned predictors that extrapolate the likely structure of the workspace from the sparse contact observations. The framework is carefully designed to handle the fact that both contact localization from joint torques and workspace prediction from sparse inputs are inherently noisy and prone to spurious hypotheses. At its core, the framework is an iterative planning and execution loop composed of three tightly coupled components:

\begin{enumerate}
    \item \textbf{Contact Detection and Isolation Module:} We employ a momentum-based observer \cite{momentumobserver1} to estimate external torques acting on the robot which allows robust detection of contact events anywhere along the manipulator. Once a contact is detected, a contact particle filter \cite{cpfmanuelli2016localizing} estimates its likely location on the robot’s surface. While binary detection is reliable, localization can be noisy, and our framework explicitly accounts for this uncertainty.  

    \item \textbf{Occupancy Estimation Module:} From the robot’s interaction history, we construct a partial estimate of the workspace: regions at estimated contact locations are marked occupied, while regions swept without collision are marked free. Because this estimate only covers explored areas, we introduce two learned predictors—a CNN and a diffusion model—that extrapolate into unexplored regions by leveraging structural priors (e.g., pipes in a sink typically span the full width). By fusing such priors with contact history, the module predicts likely occupied and free regions beyond explored space, enabling the robot to avoid potential obstacles and reach the goal efficiently.
 
    \item \textbf{Planning Module:} We integrate the workspace estimate with two existing frameworks: (i) Collision Hypothesis Sets (CHS) \cite{saundCHS} and (ii) CMAX \cite{vemula2020cmax}. This integration biases the search toward paths most likely to succeed, while preserving robustness by ensuring that feasible solutions are never discarded due to incorrect contact localization or occupancy prediction.  
\end{enumerate}

In summary, the contribution of this work is the complete framework that unifies joint torque–based contact detection and localization, occupancy prediction, and planning into a single iterative loop. We evaluate the framework in both simulation and in the real world on a UR10e manipulator across two practical domestic tasks: (i) manipulating a valve under a kitchen sink surrounded by pipes, and (ii) retrieving a target object from a cluttered shelf. Results show that the framework reliably solves these tasks, achieving up to a $2\times$ reduction in task completion time compared to baselines, with ablations confirming the contribution of each module.

\section{RELATED WORK}
Joint torque sensing has long been used for contact detection and localization by comparing measured and model-predicted torques, with the residual representing external torques due to contact~\cite{haddadin2017robot}. Momentum observers remain the standard for computing these residuals, valued for their robustness and low computational cost~\cite{momentumobserver1}. Contact localization is commonly performed with particle filters that maintain hypotheses over candidate surface points and update their likelihoods based on the observed residuals~\cite{cpfmanuelli2016localizing,zwiener2019armcl}. Our framework directly leverages these ideas for contact detection and localization. However, localization from joint torques is inherently noisy: multiple contact points can explain the same residuals, and estimates are sensitive to sensor noise and model inaccuracies~\cite{pang2021identifying}. To address this, data-driven approaches have also been explored, such as treating localization as classification over discretized surface points~\cite{popov2020transfer} or inferring contacts directly from joint kinematics without torque sensing~\cite{liang2021contact}. Such methods could serve as drop-in replacements or complements to our detection and localization module.


Unlike vision, which provides dense information, contact sensing yields sparse and local feedback, making occupancy extrapolation from structural priors essential for efficient operation. The vision community has demonstrated the power of learned priors in inferring plausible structure from incomplete data—for example, in image inpainting~\cite{lugmayr2022repaint, rombach2022high} and 3D shape completion from partial scans~\cite{dai2017shape}. At the scene level, voxel-based semantic completion models~\cite{song2017semantic} have predicted dense occupancy from single views. Our occupancy predictors can be viewed as extending these ideas to the contact-sensing realm, employing a 3D U-Net–style architecture \cite{ronneberger2015u} for the CNN-based model and the masked-resampling strategy of RePaint \cite{lugmayr2022repaint} for the diffusion model.

A number of planning frameworks for leveraging contacts in manipulation tasks have been studied. For example, \cite{saleem2023preprocessing,saleem2024pomdp} formulate the problem of localizing objects of interest through contacts as a POMDP, enabling active information gathering and robust task execution under uncertainty. \cite{javdani2013efficient} utilizes a greedy formulation for similar tasks. \cite{park2014interleaving} proposes a hierarchical planning framework for leveraging contact detections from a simulated tactile skin, while \cite{bhattacharjee2014plants} present an MPC-based framework for navigating through cluttered environments consisting of plants and tree trunks. Particularly relevant to our setting are (i) the Collision Hypothesis Set (CHS) representation \cite{saundCHS,saund2019blindfolded}, developed for planning with reliable binary contact signals, and (ii) CMAX \cite{vemula2020cmax}, a framework for planning under inaccurate models. While both are theoretically strong, they do not explicitly model or reason about the workspace, limiting their ability to exploit structural priors and often leading to repeated collisions in cluttered environments, as demonstrated by our results in Section~\ref{Section: Results}. In this work, we augment these frameworks with our workspace estimate, yielding a theoretically sound yet empirically efficient approach.

\section{Problem Setup}

Let $\mathcal{Q}$ denote the configuration space of the manipulator, and $\mathcal{W}$ be a discrete binary voxel-grid approximation of the workspace, where each voxel is either occupied by an environmental obstacle or free. Let $\mathcal{W}_{O}\subset\mathcal{W}$ be the subset of occupied voxels. The mapping $\mathcal{R}:\mathcal{Q}\to 2^{\mathcal{W}}$ returns the set of workspace voxels occupied by the robot at configuration $q\in\mathcal{Q}$; a configuration is in collision if $\mathcal{R}(q)\cap\mathcal{W}_{O}\neq\emptyset$.

Following lattice-based approaches~\cite{cohen2010search}, we cast planning as search on a graph $G=(V,E)$, where vertices $V\subset\mathcal{Q}$ are discretized configurations and edges $E$ are motion primitives from a discrete action set $\mathcal{A}$. In our case, $\mathcal{A}$ consists of linearly interpolated unit motions along each joint dimension, with transition cost $c(q,q')=\|q'-q\|$. A path $\pi=(q_0,\ldots,q_T)$ is a sequence of such edges. Executing an edge $(q_{i-1},q_i]$ either completes—certifying the swept volume $S(q_{i-1},q_i)=\bigcup_{s\in(0,1]}\mathcal{R}((1-s)q_{i-1}+s\,q_i)$ as free—or detects first contact, in which case execution halts and the robot retracts to $q_{i-1}$. We define the execution operator $\mathcal{M}(q,\pi)=(q',c)$, which returns the configuration $q'$ reached by attempting $\pi$ from $q$ and the cumulative execution cost incurred $c$.

At the start of execution, the occupancy grid $\mathcal{W}$—and hence $\mathcal{W}_{O}$—is only partially known; let the unobserved region be $\mathcal{W}_{\text{unknown}}\subset\mathcal{W}$. Proprioceptive observations $(q,\dot{q}, \tau)$ consisting of joint positions, velocities, and torques yield a reliable binary contact detection and a noisy estimate of the contact location. Contact-free traversals certify their swept volumes as free, whereas a contact indicates that $\mathcal{W}_{O}$ was intersected at some configuration $q_{\mathrm{col}}$ while executing the edge $(q_i,q_{i+1})$. The robot maintains a probabilistic occupancy grid $\hat{\mathcal{W}}_{\text{unknown}}:\mathcal{W}\to[0,1]$, where each voxel encodes its probability of being occupied as an estimate of $\mathcal{W}_{\text{unknown}}$. This estimate is refined online from contact observations and extrapolated into unexplored regions using structural priors.

Given start and goal configurations $q_{\text{start}},q_{\text{goal}}\in V$, the task is to reach $q_{\text{goal}}$ from $q_{\text{start}}$ despite partial knowledge of $\mathcal{W}_{O}$. Since collisions with unknown obstacles are unavoidable, the robot iteratively detects contacts, retracts, updates $\hat{\mathcal{W}}_{\text{unknown}}$, and replans. Within our iterative planning–execution framework, the problem can be stated as
\vspace{-0.05cm}
\begin{equation}
\begin{aligned}
\min_{\{\pi_k\}_{k=0}^{N-1}} \quad  \sum_{k=0}^{N-1} c_k \quad
\text{s.t.}\quad 
& q_0 = q_{\text{start}}, \quad q_N = q_{\text{goal}}, \\[-1em]
& (q_{k+1}, c_k) = \mathcal{M}(q_k, \pi_k).
\end{aligned}
\end{equation}

Here, $\pi_k$ denotes the path computed at the $k$-th iteration of execution. The search problem is thus defined on $G$, with edge outcomes revealed through execution. Completeness with respect to $G$ requires that if a collision-free path exists from $q_{\text{start}}$ to $q_{\text{goal}}$, the planner eventually discovers it; if none exists, infeasibility is reported in finite time. Binary contact detections are treated as reliable, while contact localizations and occupancy predictions can be noisy.

\section{METHODOLOGY}


As outlined earlier, our framework follows an iterative planning–execution loop that can be decomposed into three modules. The contact detection and localization module uses proprioceptive feedback to detect and localize contacts, while the occupancy estimation module aggregates the history of such feedback into an evolving estimate of the workspace $\hat{\mathcal{W}}_{\text{unknown}}$ and extrapolates into unexplored regions using structural priors. The planning module then carefully reasons about the reliable binary contact detections and the estimated workspace $\hat{\mathcal{W}}_{\text{unknown}}$ to generate paths that are most likely to succeed while preserving completeness.

\subsection{Contact Detection and Localization Module}
We utilize the stream of joint configurations, velocities, and torques obtained during execution to i) detect contacts and ii) localize the contact on the robot manipulator. For this purpose, we leverage established techniques that combine external momentum observers with particle filters \cite{cpfmanuelli2016localizing}.

\subsubsection{External Momentum Observer for Detection}
Following~\cite{momentumobserver1}, the generalized momentum is defined as 
$p(t) = H(q)v$, where $H(q)$ is the joint-space inertia matrix and $v$ the generalized velocity. From the manipulator dynamics, its time derivative is given by
$\dot p = \tau + \tau_{\mathrm{ext}} + C(q,v)^{\top}v - g(q)$, 
with $\tau$ the commanded joint torques, $C(q,v)$ the Coriolis and centrifugal terms, $g(q)$ gravity, and $\tau_{\mathrm{ext}}$ the external joint torques due to contact.

The momentum observer then defines a residual signal between the measured momentum and the momentum predicted by the manipulator dynamics model: 
\[
r(t) = K_O \Big(p(t) - \int_0^t \big(\tau + C(q,v)^{\top}v - g(q) + r(s)\big)\, ds \Big),
\]  
where $K_O>0$ is a diagonal gain matrix. 
This residual evolves as $\dot r(t) = K_O(\tau_{\mathrm{ext}} - r(t))$. So each component of $r$ acts as a low-pass filtered estimate of the external torque, making
the residual a good approximation of $\tau_{\mathrm{ext}}$. In nominal free-motion, $r$ remains near zero up to modeling errors and sensor noise. When an unexpected contact occurs, the external torque contribution drives the residual away from zero. A collision is declared when $|r(t)|$ exceeds a tuned threshold. The external momentum observer thus offers a lightweight, real-time method for reliable detection.

\subsubsection{Contact Particle Filter for Localization}

Once a contact has been detected, the next step is to localize its point of occurrence on the manipulator. For a given configuration $q$ and candidate location $x_c$ with Jacobian $J_{x_c}(q)$, the joint torque induced by a contact force $F_c \in \mathbb{R}^3$ is $J_{x_c}(q)^{\top}F_c$. The localization task can therefore be posed as finding the pair $(x_c, F_c)$ that best explains the observed residual $r(t) \approx \tau_{\mathrm{ext}}$. This joint optimization is non-convex due to the coupling of location and force variables. Following~\cite{cpfmanuelli2016localizing}, we approximate the solution by sampling candidate contact locations on the robot surface and solving, for each, a convex quadratic program (QP) in $F_c$, whose cost measures how well the hypothesized location explains the observed external torques.

This idea is realized in the Contact Particle Filter (CPF) \cite{cpfmanuelli2016localizing}, which maintains a set of particles $X_t = \{x_t^{[1]}, \ldots, x_t^{[M]}\}$ representing hypotheses over possible contact locations. The particle set is initialized from the set of candidate surface points $\mathcal{S}$ (i.e., $X_0 \subseteq \mathcal{S}$), and at each step is updated through resampling, perturbation, and reweighting. The measurement model assigns a likelihood to each particle based on the QP formulation. For a particle $x_t^{[m]}$ with Jacobian $J_{x_t^{[m]}}(q)$, we solve
\[
\ell(x_t^{[m]}) =
\min_{F_c \in \mathcal{F}(x_t^{[m]})}
\|\,r(t) - J_{x_t^{[m]}}(q)^{\top}F_c\,\|_{\Sigma_{\mathrm{meas}}^{-1}}^2,
\]
where $\|z\|_{A}^2 = z^{\top}Az$, $\mathcal{F}(x_t^{[m]})$ is the friction cone at $x_t^{[m]}$, and $\Sigma_{\mathrm{meas}}$ encodes measurement noise. The optimal cost is then converted into a likelihood as $p(r(t)\mid x_t^{[m]}) \propto \exp(-\tfrac{1}{2}\,\ell(x_t^{[m]}))$. Particles that yield smaller residual errors are assigned higher likelihoods, and resampling concentrates the distribution around regions of the surface consistent with the observed torques. In order to prevent particle impoverishment and stabilize the filter, a simple motion model is applied between updates, where each particle undergoes a Gaussian perturbation followed by projection back to the nearest valid point on the robot surface. After $t$ iterations, the centroid of the largest surviving cluster is returned as the contact location estimate.

\subsubsection{Robustifying Contact Localization} \label{Sec: CPF optimizations}

As noted in prior work~\cite{pang2021identifying}, multiple pairs of contact locations and forces can yield similar torque residuals, making the CPF susceptible to spurious hypotheses. To improve robustness, we incorporate a few additional refinements. First, we restrict the set of candidate points $\mathcal{S}$ to those whose motion is physically consistent with contact. Each $x \in \mathcal{S}$ has an outward surface normal $n(x)$ and an induced Cartesian velocity $\dot x(q,\dot q)$. A contact can only occur if the point is moving into the environment rather than away from it, which requires $\langle n(x), \dot x(q,\dot q) \rangle > 0$. The set of kinematically admissible candidates is therefore $\mathcal{S}_{\mathrm{active}}(q,\dot q) 
= \{x \in \mathcal{S} \mid \langle n(x), \dot x(q,\dot q)\rangle > 0\}.$
Second, we enforce workspace consistency by requiring that possible contact points also lie within unexplored regions of the workspace, i.e., $\mathcal{S}_{\mathrm{active}} \cap \hat{\mathcal{W}}_{\mathrm{unknown}}$. Finally, we also incorporate a learned link estimator: a lightweight MLP processes a temporal window of external torques $\tau_{\mathrm{ext}}$ to predict the probability of the contact having occurred at each link. These probabilities are used to lightly reweight the particles, biasing the filter toward hypotheses supported by both dynamics and the learned estimator.


\subsection{Occupancy Estimation}

The history of proprioceptive observations provides a direct but sparse estimate of the unknown workspace $\hat{\mathcal{W}}_{\text{unknown}}$. Whenever an action is executed without contact, all voxels swept by the manipulator are certified as free. Conversely, voxels corresponding to estimated contact locations are assigned high probability of occupancy. This produces a consistent map, but one that is highly incomplete: only regions the robot has explicitly interacted with are labeled. In cluttered settings such as sink cabinets or shelves, this limitation can lead to repeated collisions and inefficient exploration.

To mitigate this sparsity, we leverage structural priors of common obstacle geometries to extrapolate occupancy into unobserved regions. We cast this as a prediction problem: given a partially observed voxel grid, infer occupancy probabilities for unobserved voxels. This prediction serves as a richer estimate of $\mathcal{W}_{\text{unknown}}$ and guides subsequent planning.

We study two learned predictors for this task: a convolutional model and a generative diffusion model. Both take as input a voxel grid $\mathbf{x}\in\{0,0.5,1\}^{H\times W\times D}$, matching the dimensions of $\hat{\mathcal{W}}_{\text{unknown}}$, where $0$ denotes known free space, $1$ denotes known occupied space, and $0.5$ denotes unknown. Each predictor outputs a grid of the same dimensions, assigning to every voxel $(i,j,k)$ a continuous occupancy probability $\hat{o}_{ijk}\in[0,1]$.

\subsubsection{CNN Predictor}
Building on U-Net’s demonstrated effectiveness in vision tasks, our first predictor is a 3D U-Net~\cite{ronneberger2015u} adapted to sparse contact observations. The network follows the standard encoder–decoder design with skip connections: partial observations are processed through successive downsampling and upsampling blocks, with features passed across stages to retain spatial detail. To better capture long-range structure from limited evidence, the encoder employs large kernels with aggressive downsampling. At the end of each encoder stage, we insert a \emph{deformable sampling layer} to enrich feature extraction.

The goal of this layer is to let the network adaptively sample data-dependent features beyond the fixed voxel grid. In standard 3D deformable attention \cite{xia2022vision}, the feature at each voxel~$\mathbf{g}$ is fed into prediction heads that output a set of sampling offsets 
$\{\Delta\mathbf{g}_p\}_{p=1}^{n_\text{points}}$ and corresponding scores 
$\{s_p\}_{p=1}^{n_\text{points}}$, producing the aggregated output feature:
$\mathbf{y}(\mathbf{g})
=\sum_{p=1}^{n_\text{points}}
\operatorname{softmax}_p\bigl(s_p\bigr)\,
\mathbf{v}\bigl(\mathbf{g}+\Delta\mathbf{g}_p\bigr),
$
where $\mathbf{v}(\cdot)$ denotes the feature map from which values are sampled. However, with extremely sparse contact inputs, the softmax weights and scores can become unstable and amplify noise. To improve robustness, we keep the learnable offsets but replace the weighted sum with a max operation:
\vspace{-0.06cm}
\begin{equation}\vspace{-0.07cm}
\mathbf{y}(\mathbf{g})
=\max_{p=1,\dots,n_\text{points}}
\mathbf{v}\bigl(\mathbf{g}+\Delta\mathbf{g}_p\bigr),
\end{equation}
 which preserves adaptive sampling behavior while avoiding noisy aggregation. The decoder mirrors the encoder and outputs voxel-wise occupancy probabilities $\hat{o}\in[0,1]$ through a final sigmoid layer, which we interpret as $\hat{\mathcal{W}}_{\text{unknown}}$. Training uses standard binary cross-entropy loss augmented with a per-voxel binary entropy regularizer to discourage over-confident predictions.

\subsubsection{Diffusion Predictor}

To capture the multi-modality inherent in extrapolating from sparse contacts, we also develop a generative predictor based on diffusion models. We adapt the same U-Net backbone as above, conditioning on the partial observation $\mathbf{x}^{\text{obs}}$ and a binary mask $m$ marking observed ($m=1$) and unobserved ($m=0$) voxels. For inference, we apply DPM-Solver++~\cite{lu2025dpm}, a high-order ODE sampler that enables fast and stable generation.


A key challenge is generating occupancy grids consistent with the observed regions provided as part of the input. Standard diffusion inpainting gradually denoises the entire grid, which can drift even in known voxels. To prevent this, we adopt a \emph{known-region reinforcement} strategy inspired by RePaint~\cite{lugmayr2022repaint}: after each denoising step, the sample $\mathbf{x}_t$ is updated as $\mathbf{x}_t \leftarrow m\odot \mathbf{x}^{\text{obs}}_{t} + (1-m)\odot \mathbf{x}_t,$ where $\mathbf{x}^{\text{obs}}_{t}$ is the noisy version of the observed input. This reinsertion biases the predictions to be consistent with the partial inputs.

We further apply classifier-free guidance~\cite{ho2022classifier}, which biases the denoising trajectory toward the given condition $c$ (in our case, the partial input) by combining conditional and unconditional predictions:
\[
\hat{\epsilon}_\theta=\epsilon_\theta(\mathbf{x}_t,t,\varnothing, \mathbf{m})+
w\big(\epsilon_\theta(\mathbf{x}_t,t,\mathbf{x}^{\text{obs}},\mathbf{m})
-\epsilon_\theta(\mathbf{x}_t,t,\varnothing, \mathbf{m})\big),
\]
where $w>1$ controls the strength of conditioning. This balances two objectives: enforcing consistency with observed voxels while still exploring diverse and plausible completions in unobserved regions. Unlike the CNN predictor, the diffusion model represents the full conditional distribution $p(\mathbf{x}\mid \mathbf{x}^{\text{obs}})$. We approximate the expected occupancy volume $\hat{\mathcal{W}}_{\text{unknown}}$ by generating a batch of samples and taking their voxel-wise mean.

\subsection{Planning Module}

Given the observation history and the estimated occupancy $\hat{\mathcal{W}}_{\text{unknown}}$, the planning module computes paths that maximize success probability—avoiding collisions while reaching the goal with minimum execution cost. To guarantee completeness, it must avoid pruning feasible solutions due to noisy predictions while preventing indefinite execution of failing plans. To achieve this, we propose two variants that combine the estimated occupancy map with two previously developed ideas:  
(i) \emph{Collision Hypothesis Sets} (CHS) \cite{saundCHS}, which encode reliable binary contact detections into a sparse set-based representation; and  
(ii) \emph{CMAX}~\cite{vemula2020cmax}, a framework for planning with inaccurate world models.  

\subsubsection{Integration with CHS}
\begin{figure}
\centering
  \includegraphics[width=0.4 \textwidth]{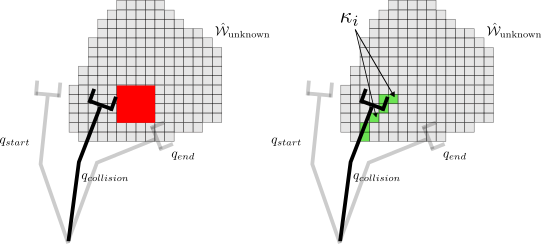}
  \caption{The collision hypothesis set $\kappa_i$ (in green) corresponds to the voxels intersecting with the active surface of the robot at the time of collision.} \label{Fig: CHS}
  \vspace{-1em} \vspace{-0.2cm}
\end{figure}

A single CHS $\kappa_i \subset \mathcal{W}$ is the set of voxels that could explain a collision observed when executing an edge $e = (q_i, q_{i+1})$. If a collision is detected at configuration $q_{\mathrm{col}}$, then among the voxels intersecting the active robot surface, at least one must be occupied (refer Fig.~\ref{Fig: CHS}); this set of potentially occupied voxels is recorded as $\kappa_i$. The CHS representation maintains the collection $\kappa = \{\kappa_1, \kappa_2, \ldots \}$ of all such hypotheses accumulated during execution.

Given a hypothesis $\kappa_i$, the probability that an edge $e$ is in collision is defined in proportion to the fraction of the hypothesis set swept by the edge:
\begin{equation}
    P(v(e) = 0 \mid \kappa_i) = \frac{|S(e) \cap \kappa_i|}{|\kappa_i|}.
\end{equation}
If $S(e)$ fully contains $\kappa_i$, then $P(v(e) = 0 \mid \kappa_i) = 1$, since by construction $\kappa_i$ must include at least one occupied voxel. Assuming independence across hypotheses, the probability that $e$ is valid given all CHSs is then given by:
\begin{equation} \label{Eqn: CHS cost} \vspace{-0.15cm}
    P(v(e) = 1 \mid \kappa) = \prod_i \big(1 - P(v(e) = 0 \mid \kappa_i)\big).
\end{equation}
Edges that have already been attempted and invalidated receive zero validity probability, ensuring they are never re-executed, while other edges are never prematurely ruled out.

While CHS is conservative and accurate, it does not explicitly construct the workspace, which limits the ability to exploit structural priors, often resulting in many contacts and poor performance in cluttered regions. We therefore augment it with $\hat{\mathcal{W}}_{\text{unknown}}$, biasing search toward safer trajectories. Concretely, we perform a search on the graph $\mathcal{G}$ with a modified edge cost function. Each edge cost is augmented with penalties derived from both CHS and the occupancy map, reflecting the likelihood that the edge leads to a collision under each representation. Edges assigned infinite cost are deemed infeasible and never selected for execution. Formally, for each edge $e$, the cost is defined as
\begin{equation} \label{Eqn: cost function}
    c(e) = c_{\text{dist}}(e) + \alpha \, c_{\text{CHS}}(e) + \beta \, c_{\hat{\mathcal{W}}_{\text{unknown}}}(e),
\end{equation}
where $c_{\text{dist}}(e)$ is the configuration-space distance, 
$c_{\text{CHS}}(e) \propto \tfrac{1}{P(x(e) = 1 \mid \kappa)}$ penalizes edges with low CHS validity, and $c_{\hat{\mathcal{W}}_{\text{unknown}}}(e)$ penalizes edges likely to collide according to the occupancy estimate. In practice, $\alpha \gg \beta$ as CHS is sparse but highly reliable and can prune edges outright (edges with zero validity probability according to Eqn. \ref{Eqn: CHS cost} incur infinite cost), whereas occupancy-based penalties are bounded and used only to bias the search.


\subsubsection{Integration with CMAX}

CMAX~\cite{vemula2020cmax} addresses planning with inaccurate models by interleaving planning and execution while preserving real-time operation and completeness. When execution reveals that a transition differs from the nominal model, CMAX does not repair the model but instead biases the planner away from such transitions by inflating their cost—typically by placing hyperspheres around discovered discrepancies so that nearby state–action pairs incur additional penalty. As new discrepancies are observed, the process repeats, ensuring the planner ultimately avoids all incorrect transitions and still reaches the goal.  

In our setting, the nominal model assumes the workspace is empty, so collisions reveal discrepancies between $\hat f$ and the true dynamics $f$. Let $\chi \subseteq S \times A$ denote the set of state–action pairs producing such discrepancies. The CMAX penalty for a candidate $(s,a)$ is defined as  
\vspace{-0.1cm}
\begin{equation} \vspace{-0.08cm}
c_{\text{CMAX}}(s,a)\;\propto\;
\max_{\substack{(s',a') \in \chi,\, a' = a \\ d(s,s') < \delta}}
\frac{1}{d(s,s')},
\end{equation}
where $d(\cdot,\cdot)$ is the Euclidean distance in configuration space and $\delta$ is a domain-dependent hypersphere radius. Transitions that repeat action $a$ near a previously invalidated state $s'$ incur large penalties, diverging as $d(s,s') \to 0$, while those outside all $\delta$-neighborhoods remain unpenalized. Although theoretically sound, this framework by itself can be less effective in our setting, since penalization is applied in the state–action space. Consequently, the robot can execute different transitions that appear unrelated in state–action terms yet traverse overlapping workspace regions, resulting in repeated collisions.

To mitigate this limitation, we augment the CMAX penalty with workspace-level costs derived from our occupancy estimate, similar to the CHS case. Hence, similar to Eqn. \ref{Eqn: cost function}, the cost of an edge is defined as the linear sum of the baseline path length, the CMAX penalties, and the occupancy-based penalties $c_{\hat{\mathcal{W}}_{\text{unknown}}}(e)$. As in the CHS case, CMAX can prune edges outright by assigning infinite cost to known-invalid transitions, whereas occupancy-based penalties are bounded and serve only to bias the search. 

\paragraph*{Completeness Discussion} 
Both the CHS representation and the CMAX framework are theoretically sound and complete, guaranteeing recovery of a path in $\mathcal{G}$ (if one exists) when binary contact signals are used to identify infeasible edges. They operate through distinct mechanisms: CHS constructs a hypothesis set $\kappa_i$ containing at least one truly occupied voxel and permanently invalidates edges whose swept volume fully covers $\kappa_i$, whereas CMAX records discrepancies and inflates the costs of nearby state–action pairs, assigning infinite cost only to the invalidated transition itself. In both cases, only edges that are provably infeasible are removed, while all other edges remain in the search space; hence, a feasible edge is never falsely discarded. Since the estimated workspace is used solely to bias costs and never to prune, all feasible edges in $\mathcal{G}$ remain admissible, as guaranteed by the vanilla variants. Consequently, the search will eventually identify a feasible path if one exists—though it may first explore paths that appear promising under $\hat{\mathcal{W}}_{\text{unknown}}$ and then backtrack when they are invalidated— establishing completeness with respect to $\mathcal{G}$.

\section{RESULTS} \label{Section: Results}
\begin{figure*}
\centering
  \includegraphics[width=0.95\textwidth]{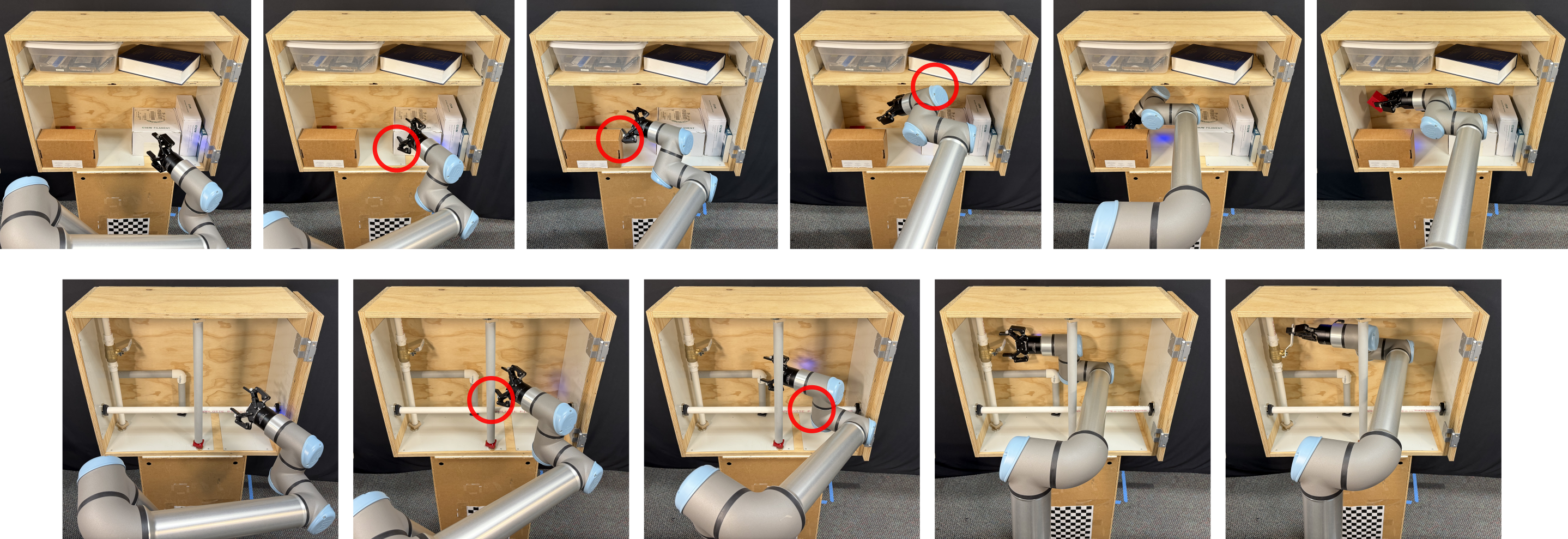}
  \caption{Example runs of the framework on the object retrieval task (top) and valve manipulation task (bottom). In both cases, the robot initially collides with environmental obstacles (highlighted in red) before identifying a feasible path that allows it to successfully reach the target.
}\label{fig:manipulation_real_robot}\vspace{-0.35cm} 
\end{figure*}

We evaluate the proposed framework using a UR10e manipulator in two representative domains that capture the challenges of manipulating in the blind. The first domain involves \emph{valve manipulation under a sink cabinet}, where the robot must reach a shutoff valve mounted at the back wall while maneuvering around pipes that span across the cavity. This setting emphasizes strong structural priors, as pipes are slender and typically extend across the full width of the cabinet. The second domain involves \emph{object retrieval from a cluttered bookshelf}, where the robot must reach deep into a shelf to retrieve an item placed at the back while interacting with books, shelf partitions, and other irregularly shaped objects (Fig.~\ref{fig:manipulation_real_robot}). 

To illustrate the impact of the different modules, we evaluate the following planner variants. 
\textbf{CHS} and \textbf{CMAX} are the vanilla frameworks that do not utilize the estimated workspace occupancy $\hat{\mathcal{W}}_{\text{unknown}}$. 
\textbf{CHS + CNN/Diffusion} and \textbf{CMAX + CNN/Diffusion} are the proposed extensions that incorporate the estimated occupancy, obtained from either a CNN- or diffusion-based predictor. Finally, we consider variants that bypass both CHS and CMAX and instead plan directly on the estimated occupancy map $\hat{\mathcal{W}}_{\text{unknown}}$. In these cases, a voxel is treated as occupied if its estimated occupancy exceeds a fixed threshold (0.8), and any edge intersecting such voxels is ruled infeasible. Similar to Eqn.~\ref{Eqn: cost function}, edge costs here consist of configuration distance and occupancy penalties. These variants are denoted as \textbf{$\hat{\mathcal{W}}_{\text{unknown}}$ (no pred)}, and \textbf{$\hat{\mathcal{W}}_{\text{unknown}}$ + CNN/Diffusion}, depending on which predictor they employ.

We evaluate the frameworks in both simulation and real-world across the two domains. All planners used Weighted A* search with a planning timeout of 5\,s per iteration. A trial was declared a failure if (i) no solution was found within the timeout or (ii) the framework exceeded 20 plan--execute iterations without reaching the goal. In the reported tables, ``Num iters'' denotes the number of plan--execute iterations required to solve a problem, where each iteration consists of executing the planned motion, estimating contacts, updating $\hat{\mathcal{W}}_{\text{unknown}}$, and replanning. Planning, execution, occupancy prediction, and contact-estimation times are reported as cumulative time spent on each component across all iterations. The ``Total time'' entry corresponds to the overall time required by each variant to complete the task. 

In both domains, the unknown portion of the workspace ${\mathcal{W}}_{\text{unknown}}$ corresponds to a tightly constrained cabinet with multiple potential obstacles. In the pipe domain, ${\mathcal{W}}_{\text{unknown}}$ contains 5--12 randomly placed pipe segments inside the cabinet, while in the shelf domain, ${\mathcal{W}}_{\text{unknown}}$ contains 1--4 shelf partitions and 6--12 randomly placed household objects. The contact estimation module was used only in the real-world studies; in simulation, contact estimation was modeled by perturbing the true collision point in Cartesian space with zero-mean Gaussian noise (standard deviation $3$\,cm independently along each coordinate) before projecting it back onto the robot surface as the estimated contact location.

\paragraph*{Training Data for Occupancy Predictor} Training data for the occupancy prediction modules were generated by sampling random actions and recording interactions with environmental obstacles in both domains. For each datapoint, a random scene was sampled, and 1–7 actions were executed. The input is the partial occupancy map derived from these sweeps. The output has voxels corresponding to objects the robot directly interacted with marked occupied, while all other voxels are labeled free. To simplify learning and reduce ambiguity, supervision is restricted to these interacted objects rather than the entire scene. This design choice reflects the limited spatial correlation between objects in our environments and yields a more tractable prediction problem.

\subsection{Simulation Results}

\begin{table*}[t]
\centering
\caption{Comparison of planners in simulation for valve manipulation (pipes) and object retrieval (shelf) tasks.} \label{Table: Sim Results}
\setlength{\tabcolsep}{7.5pt}
\begin{tabular}{lcccccc|cccccc} 
\toprule
& \multicolumn{6}{c}{\textbf{Pipes (Valve Manipulation)}} 
& \multicolumn{6}{c}{\textbf{Shelf (Object Retrieval)}}\\
\cmidrule(lr){2-7}\cmidrule(lr){8-13}
\textbf{Planner} &
\makecell{Succ.\\Rate} &
\makecell{Num\\Iter.} &
\makecell{Plan. \\ (s)} &
\makecell{Exec. \\ (s)} &
\makecell{Pred. \\(s)} &
\makecell{Total \\(s)} &
\makecell{Succ.\\Rate} &
\makecell{Num \\ Iter.} &
\makecell{Plan. \\ (s)} &
\makecell{Exec. \\(s) } &
\makecell{Pred. \\(s)} &
\makecell{Total \\(s)}\\
\midrule
CHS & 0.83 & 11.15 & 0.48 & 346.66 & 0 & 347.14 & 0.52 & 10.69 & 12.55 & 309.47 & 0 & 322.02\\
CHS + CNN & \bfseries1.00 & 5.76 & 2.41 & 195.97 & 1.84 & 200.22 & \bfseries0.94 & 4.20 & 11.22 & 135.56 & 1.55 & \bfseries148.34\\
CHS + Diffusion & 0.97 & 4.95 & 1.52 & 191.94 & 97.64 & 291.10 & 0.91 & 4.15 & 13.39 & 141.98 & 80.1 & 240.53\\
\midrule
CMAX & 0.57 & 13.14 & 16.03 & 408.29 & 0 & 424.31 & 0.33 & 13.88 & 12.38 & 380.32 & 0 & 392.69\\
CMAX + CNN & 0.86 & 6.14 & 11.48 & 237.82 & 1.96 & 251.26 & 0.73 & 4.11 & 9.91 & 141.47 & 1.56 & 152.95\\
CMAX + Diffusion & 0.78 & 5.93 & 11.63 & 184.34 & 116.96 & 312.94 & 0.66 & 5.67 & 15.5 & 180.9 & 100.93 & 303.66\\
\midrule
$\hat{W}_{\text{unknown}}$ (no pred) & 0.59 & 10.95 & 0.13 & 333.57 & 0 & 333.70 & 0.47 & 12.83 & 18.54 & 346.62 & 0 & 365.16\\
$\hat{W}_{\text{unknown}}$ + CNN & 0.82 & 5.86 & 1.51 & 192.02 & 1.87 & \bfseries195.41 & 0.89 & 5.08 & 14.20 & 168.53 & 1.42 & 184.15\\
$\hat{W}_{\text{unknown}}$ + Diffusion & 0.79 & 5.04 & 1.46 & 187.82 & 99.30 & 288.59 & 0.81 & 4.81 & 16.32 & 159.97 & 89.34 & 271.25\\
\bottomrule
\end{tabular} \vspace{-0.4cm}
\end{table*}

Table~\ref{Table: Sim Results} summarizes simulation results for both domains, averaged over 200 randomly generated problem instances. Overall, the \textbf{CHS + CNN} variant achieved the strongest performance, solving 100\% of all sampled problems. This variant effectively combines the strengths of both modules: CNN-based predictions bias the search away from repeated collisions, while the CHS representation ensures robustness by falling back to reliable binary detections when predictions are noisy. By contrast, \textbf{CHS alone} achieves lower success rates, as the robot often requires many more interactions with the environment before reaching the goal. On average, the total time required by CHS + CNN is nearly half that of CHS, confirming the empirical speedup provided by the occupancy estimation and prediction module. Similarly, planning directly with just the workspace estimate $\hat{\mathcal{W}}_{\text{unknown}}$ (without CHS or CMAX) yields lower success rates due to false occupancy estimations causing the planner to incorrectly declare infeasibility when a solution exists. Combining the two approaches, therefore provides the best of both worlds—leaning on the occupancy estimator when predictions are accurate, while falling back on binary detections when they are not. 

Between the two base planners, CHS consistently outperforms CMAX. Although CHS does not explicitly build a workspace estimate, its hypothesis-set representation still captures portions of the workspace that are potentially occupied, enabling it to avoid redundant collisions. CMAX, by contrast, penalizes discrepancies only in state--action space and therefore suffers from repeated collisions with obstacles. When augmented with occupancy prediction, however, CMAX shows clear improvements: \textbf{CMAX + CNN} narrows the performance gap to CHS, demonstrating the utility of workspace-level reasoning. Across both base frameworks, adding a predictor reduces the number of plan--execute iterations by more than half, directly translating into lower execution times and faster task completion.

A domain-level comparison shows that shelf environments are more challenging: baseline planners without predictors (CHS, CMAX, and $\hat{\mathcal{W}}_{\text{unknown}}$) achieve lower success rates and higher planning times on shelves than on pipes due to heavier clutter and more complex maneuvers required. However, when combined with prediction, success rates in shelves match or even exceed those in pipes, since predictors can exploit the structured regularities of shelves more effectively than randomized pipe routings. On average, predictor-augmented frameworks reduce total task time by a factor of 1.5--2$\times$, with improvement particularly pronounced in shelf domains where structural priors are strongest.

Finally, we note the trade-off between CNN and diffusion predictors. Both reduce iterations and execution time to a similar extent, confirming comparable prediction quality. However, diffusion-based predictors are computationally intensive: generating each occupancy update requires sampling (50 samples in our implementation) and voxelwise averaging, incurring an additional $\sim$20\,s per prediction. In contrast, the CNN model provides fast inference ($\sim$0.3s per prediction) while achieving equivalent planning and execution gains. As a result, CNN-based prediction offers the most practical balance between accuracy and efficiency, while diffusion provides a proof of concept for handling multi-modality at the cost of higher runtime.

\subsection{Real Robot Results}


\begin{table}[t]
\centering
\caption{Real-world valve manipulation and object retrieval results} \label{tab:real_results}
\setlength{\tabcolsep}{5pt}
\begin{tabular}{lcccccc}
\toprule
\textbf{Planner} &
\makecell{Num\\Iter.} &
\makecell{Plan.\\(s)} &
\makecell{Exec.\\(s)} &
\makecell{Pred.\\(s)} &
\makecell{Contact\\Est. (s)} &
\makecell{Total\\(s)} \\
\midrule
\multicolumn{7}{c}{\textbf{Pipes (Valve Manipulation)}}\\
\midrule
$\hat{W}_{\text{unknown}}$ & 10.7 & 0.09 & 324.2 & 0 & 21.8 & 357.1 \\
$\hat{W}_{\text{unknown}}$ + CNN     & 5.0  & 5.17 & 192.8 & 1.61 & 10.7 & 214.5 \\
CHS                                   & 11.0 & 0.25 & 324.1 & 0 & 24.7 & 360.4 \\
CHS + CNN                             & 3.8  & 1.44 & 161.2 & 1.11 & 8.9 & 175.9 \\
\midrule
\multicolumn{7}{c}{\textbf{Shelf (Object Retrieval)}}\\
\midrule
$\hat{W}_{\text{unknown}}$  & 13.3 & 20.6 & 404.5 & 0 & 26.0 & 466.0 \\
$\hat{W}_{\text{unknown}}$ + CNN     & 5.7  & 12.7 & 168.5 & 1.71 & 11.4 & 199.8 \\
CHS                                   & 12.6 & 19.8 & 344.7 & 0 & 27.9 & 407.5 \\
CHS + CNN                             & 4.1  & 11.3 & 110.6 & 1.60 & 10.8 & 138.0 \\
\bottomrule
\end{tabular} \vspace{-0.55cm}
\end{table}

We further validated the framework on a UR10e manipulator in the real world across both domains, using the best-performing variants from simulation. Results are averaged over 20 runs per domain. A key difference from simulation is the use of raw proprioceptive signals for contact estimation: joint currents (linearly proportional to joint torques) were monitored to estimate external torques $\tau_{\text{ext}}$, and the contact particle filter was run with 250 particles per update.

The performance trends observed in simulation largely carried over to hardware (Table~\ref{tab:real_results}). In particular, the \textbf{CHS + CNN} variant consistently outperformed all others, reducing execution time to less than half of vanilla CHS. For example, in the shelf domain, CHS + CNN reduced the total task time from $407.5$\,s (CHS) to $138.0$\,s. Similarly, in the pipe domain, CHS + CNN completed tasks in $175.9$\,s compared to $360.4$\,s for CHS. These results demonstrate that the benefits of occupancy prediction are not limited to simulation but transfer effectively to real-world execution.


Assuming a localization error of under 4cm as successful, the contact detection and localization module achieved $73\%$ accuracy with the optimizations from Section~\ref{Sec: CPF optimizations} (Robustifying CPF) enabled. An ablation study on a held-out dataset attributed a $20\%$ accuracy gain to these optimizations compared to a baseline without them. Although localization remains noisy, the CHS abstraction ensures that binary detections provide a reliable fallback when predictions are incorrect.

The benefits of occupancy prediction were most pronounced in the shelf domain, where naive CHS and $\hat{\mathcal{W}}_{\text{unknown}}$ variants often required repeated probing to escape occluded regions. By leveraging predicted occupancy, the robot was able to anticipate unseen obstacles and reduce redundant collisions. Importantly, the CHS representation provided robustness against prediction errors: when contact was mislocalized, CHS still guided the planner away from the true contact region, allowing the robot to progress without re-colliding. In contrast, the $\hat{\mathcal{W}}_{\text{unknown}}$ + CNN variant typically required one or two additional iterations to revisit the same obstacle before updating its estimate. This explains the performance gap between $\hat{\mathcal{W}}_{\text{unknown}}$ + CNN and CHS + CNN, despite both using the same occupancy predictor.

\section{FUTURE WORK AND CONCLUSION}
A promising future direction is to augment occupancy prediction with textual priors. Instead of conditioning only on a partial occupancy map, the predictor could also take natural language prompts (e.g., “kitchen sink with pipes” or “bookshelf with partitions”). As proof of concept, we extended our CNN predictor with CLIP-based prompt embeddings~\cite{radford2021learning}, injected via FiLM modules~\cite{perez2018film} at each encoder stage. This text conditioning enabled the CNN to bias its predictions toward the described domain and, when integrated with the planner, the model performed competitively with the baseline CNN, achieving average task completion times of $216$ seconds (pipes) and $157$ seconds (shelves) in simulation. These results highlight the potential of textual priors. Further details will appear in the extended version.

In conclusion, we introduced a theoretically complete and empirically efficient framework for manipulating using contacts. The framework tightly couples three components—torque-based contact detection and localization, workspace occupancy estimation, and planning—while carefully reasoning about noise and uncertainty in both contact localization and occupancy prediction. Experiments in both simulation and in the real world show that our framework reliably completes tasks such as valve manipulation and object retrieval, achieving up to a $2\times$ reduction in task completion time compared to baselines. 
\bibliographystyle{plain}
\bibliography{references}

\end{document}